\documentclass{article}

\usepackage[preprint]{neurips_2020}

\usepackage[utf8]{inputenc} %
\usepackage[T1]{fontenc}    %
\usepackage{hyperref}       %
\usepackage{url}            %
\usepackage{booktabs}       %
\usepackage{amsfonts}       %
\usepackage{nicefrac}       %
\usepackage{microtype}      %
\usepackage{xcolor}

\usepackage{amsmath}
\DeclareMathOperator*{\maxpool}{maxpool}

\usepackage[pdftex]{graphicx}   %
\usepackage{subcaption}		    %
\usepackage{wrapfig}            %
\usepackage{enumitem}           %

\usepackage{amsthm}
\theoremstyle{plain}

\theoremstyle{remark}

\theoremstyle{definition}

\renewcommand{\t}[1]{\texttt{#1}}

\setitemize{noitemsep,topsep=0pt,parsep=0pt,partopsep=0pt,leftmargin=*}
\setenumerate{noitemsep,topsep=0pt,parsep=0pt,partopsep=0pt,leftmargin=*}

\title{Neural Program Synthesis with a Differentiable Fixer}

\author{%
  Matej Balog\thanks{Work done during an internship at Google Research, Brain Team.} \\
  University of Cambridge\\
  \& MPI Tübingen
  \And
  Rishabh Singh \\
  Google Research \\
  Brain Team \\
  \And
  Petros Maniatis \\
  Google Research \\
  Brain Team \\
  \And
  Charles Sutton \\
  Google Research \\
  Brain Team \\
}

\begin{document}

\maketitle

\begin{abstract}
We present a new program synthesis approach that combines an encoder-decoder based synthesis architecture with a differentiable program fixer. Our approach is inspired from the fact that human developers seldom get their program correct on the first attempt, and perform iterative testing-based program fixing to get to the desired program functionality. Similarly, our approach first learns a distribution over programs conditioned on an encoding of a set of input-output examples, and then iteratively performs fix operations using the differentiable fixer. The fixer takes as input the original examples and the current program's outputs on example inputs, and generates a new distribution over the programs with the goal of reducing the discrepancies between the current program outputs and the desired example outputs. We train our architecture end-to-end on the RobustFill domain, and show that the addition of the fixer module leads to a significant improvement on synthesis accuracy compared to using beam search.
\end{abstract}

\section{Introduction}
\label{sec:introduction}

Program synthesis, the task of automatically generating programs that satisfy a given specification~\citep{sygus,now17}, is a longstanding fundamental problem in artificial intelligence~\citep{mannaw71,summers1977methodology}. 
Perhaps the simplest way to specify what a program should do is to provide a few
examples of program inputs and desired outputs, a paradigm called \emph{programming by example (PBE)}~\citep{lieberman2001your,flashfill}.
PBE synthesis systems have been developed for various domains, including string transformations~\citep{flashfill,devlin2017robustfill}, list manipulations~\citep{balog2017deepcoder}, and CAD images~\citep{ellis2019repl}.

Human programmers, or at least the authors, seldom write a totally correct program on the first attempt.
Instead, we write some code, 
test it on a set of examples, 
and then iteratively fix the code until it produces the desired outputs. Current machine-learning approaches for PBE do not do this, however.
They search over programs guided by a learnt distribution~\citep{devlin2017robustfill,balog2017deepcoder}, but they do not
learn how to fix complete predicted programs that turn out to be incorrect.

We present an architecture for neural program synthesis based on a
\emph{differentiable fixer} that is trained to make changes to complete but incorrect
programs. The fixer works on top of a base neural program synthesizer based
on sequence-to-sequence learning \citep{devlin2017robustfill}. 
First, the program predicted by the base synthesizer is executed on the inputs
to produce predicted outputs.
If these outputs do not equal the correct outputs,
the differentiable fixer generates a new program based on the inputs, predicted outputs, and correct outputs. If the prediction from the fixer is still incorrect,
this procedure can  be iterated until either the predicted program generates the desired output or we reach the maximum number of fix steps.
Because the fixer is differentiable, we can train it end-to-end alongside the base neural synthesizer.

The differentiable fixer is related to execution-guided synthesis methods 
\citep{zohar2018pccoder, ellis2019repl, egnps}, which execute many partial programs, using 
the intermediate results to guide the search.
This idea is powerful, but it places limitations on the language and search techniques used, because the partial programs must be executable.
The differentiable fixer is a complementary idea, because it can make changes, at potentially arbitrary locations, to a complete program. 

We instantiate the differentiable fixer in the domain of RobustFill~\citep{devlin2017robustfill},
which defines a language of string transformations that commonly occur
in spreadsheet formulas. 
Compared to a RobustFill baseline that makes multiple predictions using beam search,
the fixer achieves higher accuracy. We also perform several ablation studies to better understand the qualitative differences. Interestingly,
we find that the fixer makes many more changes to incorrect programs than beam search,
and the fixer is more likely to make changes earlier in the program.
This supports the hypothesis that the fixer is producing a more holistic representation of how to fix the program.

\vspace{-0.5em}
\section{Related work}
\label{sec:background}
\vspace{-0.5em}

In this work we are concerned with \emph{program synthesis}, conventionally formulated as a combinatorial search problem over the space of programs with the goal of finding a program matching a given specification.
Recently, \emph{neural program synthesis}~\citep{parisotto2017neuro} techniques have been proposed to accelerate the search process by training on a large amount of synthetically generated (specification, program) pairs. RobustFill~\citep{devlin2017robustfill} takes an end-to-end approach, with the neural network consuming the specification (here a set of input-output examples), and predicting the target source code token-by-token.
Alternative methods include SketchAdapt~\citep{nye2019sketchadapt}, which predicts 
a rough sketch of the program with holes to be filled in using symbolic search techniques, and DeepCoder~\citep{balog2017deepcoder}, which predicts properties of the target source code that are then used to guide a conventional synthesizer.
In DeepCoder, the neural network predicitions are passed as hints in the beginning of the search, but the network is not queried again.

When the neural network is embedded within the search procedure, the network can also take as input information about the partial solution constructed so far (in addition to the original program specification).
In \emph{execution-guided program synthesis}~\citep{zohar2018pccoder, egnps, ellis2019repl} this additional input comprises the result of executing the partial program constructed by the search so far on the inputs in the input-output specification.
This permits the search to proceed in semantic space~\citep{ellis2019repl}, where the network guides the iterative extension of partial programs until one matching an input-output specification is found.

Our work is closely related in spirit to execution-guided program synthesis but fundamentally different in its underlying assumptions:
(1) we do not rely on a domain-specific language (DSL) where partial programs are executable, nor do we require that the learned component builds the program bottom-up;
(2) we are able to apply the differentiable fixer to incorrect predictions iteratively until a valid solution is found, whereas execution-guided approaches need to restart (with a larger number of particles, say) if none of the initial predictions work.
Moreover, our technique is able to fix initially incorrect programs in arbitrary locations of the source code, not just at the location that is currently being constructed by a bottom-up execution-guided search procedure.
That said, execution-guided synthesis is a powerful technique when it is applicable, and can be combined with our complementary technique of fixing incorrect full predictions.

Our work can be also interpreted as jointly learning a latent continuous representation of programs and a fixer module that utilizes execution information from an initially predicted program to propose a fixed representation in this latent space.
Learning continuous representations for discrete objects has been a longstanding goal of representation learning, as such a continuous space can facilitate differentiable reasoning~\citep{lee2019reasoning}, optimization of, and discovery of new discrete objects with desirable properties~\citep{bombarelli2018molecules}.
Continuous representations have been learned, e.g., for computer programs~\citep{piech2015student}, symbolic expressions~\citep{allamanis2017eqnet}, logical formulas~\citep{evans2018entailment}, and molecules~\citep{bombarelli2018molecules}.

There are also some recent related works on specification-guided \emph{neural program repair}~\citep{shin2018towards, nscorrector, skp, dynamicrepair}, where a network is trained to repair a candidate program based on a desired specification of I/O examples or functional behavior. Unlike our work, these approaches for program repair do not attempt to solve the synthesis problem (i.e. generate a program from scratch), because a (buggy) human-written program is given as input. 

Finally, 
\emph{program induction} learns a program that is represented implicitly in the weights of a neural network mapping from program inputs to program outputs~\citep{graves2014neural, reed2016neural}. This is a separate problem from {program synthesis}, which aims
to explicitly generate source code in a programming language~\citep{hardy1975example, menon2013example}.

\section{Synthesis with a Differentiable Fixer}
\label{sec:method}

In this section, we first present a brief overview of the synthesis problem formulation and the baseline RobustFill architecture. We then present our new differentiable fixer architecture and describe some of the key design choices in designing and combining the fixer with the synthesis architecture.

\textbf{Preliminaries and Notation.}
In a domain-specific language (DSL) $\mathcal{L}$, our goal is to learn a program synthesizer $\mathcal{S}$ that, given a set of input-output examples $E = \{(i_1, o_1), \cdots, (i_N, o_N)\}$, generates a program $P \in \mathcal{L}$ whose semantics is consistent with the examples: $\forall (i_n, o_n) \in E: P(i_n) = o_n$.

The DSL for string transformations $\mathcal{L}$ that we consider in this work is shown in Figure~\ref{fig:dslexample}(a); this DSL is similar to those of NSPS~\citep{parisotto2017neuro} and RobustFill~\citep{devlin2017robustfill}.
The top-level expression in the DSL is a \t{Concat} operator that concatenates the results of its argument expressions.
Each argument expression can either be a constant string \t{s} or a substring expression on the input \t{i} that takes two position-expression arguments for left and right indices, respectively.
A position expression can be a regular expression \t{Regex} or a constant position \t{ConstPos}.
A regex position expression $\t{Regex}(r, k, \t{Start})$ returns the starting position of the $k^{\text{th}}$ match of the regular expression $r$; the position expression for the end of a match is defined similarly.
Negative values for $k$ and $n$ indicate counting (matches or positions, respectively) from the end of the input string. A small set of regular expressions are allowed, including
expressions that match alphabetic words, sequences of digits, and so on.
Similarly, a small vocabulary of generally-useful constant strings are allowed, such as common
punctuations.
An example synthesis task with four input-output examples and one of the possible programs in the DSL that conforms to the examples is shown in Figure~\ref{fig:dslexample}(b).

\begin{figure}[!tb]
\begin{tabular}{c|c}
\begin{minipage}{0.4\linewidth}
\begin{center}
{\small
\begin{eqnarray*}
P & := & \t{Concat}(e_1, \cdots, e_n) \nonumber \\
e & := & \t{ConstStr}(s) \\
& | & \t{SubStr}(p_1, p_2) \nonumber \\
p & := & \t{Regex}(r, k, \t{Start}) \\
& | & \t{Regex}(r, k, \t{End}) \; | \; \t{ConstPos}(n) \nonumber \\
r & := & s \; | \; t \nonumber \\
n & := & -L \; | \; \cdots \; | \; -1 \; | 0 \; | \; 1 \; | \; \cdots \; | \; L \nonumber \\
k & := & -K \; | \; \cdots \; | \; -1 \; | 0 \; | \; 1 \; | \; \cdots \; | \; K \nonumber \\
t & := & \t{Word} \; | \; \t{Num} \; | \cdots | \; \t{Digits} \nonumber \\
s & := & \t{" "} \; | \; \t{"."} \; |\; \t{". "} \; | \cdots | \; \t{","} \nonumber \\
\end{eqnarray*}
}
\end{center}
\end{minipage}

&

\begin{minipage}{0.6\linewidth}
\footnotesize
\begin{tabular}{|c|c|c|}
\hline
& \multicolumn{1}{|c|}{\bf Input (i)} & \multicolumn{1}{|c|}{\bf Output (o)} \\\hline\hline
1 & Mark Henry, 521-625-2716 & {M. Henry : 521} \\ \hline
2 & Barry M. Myers, 617-278-8787 & {B. Myers : 617} \\ \hline
3 & Michael Jones, 425-267-2871 & {M. Jones : 425} \\ \hline
4 & Jon Sanders, 617-225-9819 & {J. Sanders : 617} \\ \hline
\end{tabular}
{\footnotesize
\begin{verbatim}
Concat(SubStr(ConstPos(0), ConstPos(1)),
       ConstStr(". "),
       SubStr(Regex(Word, -1, Start),
              Regex(",", 1, Start)),
       ConstStr(" : "),
       SubStr(Regex(Num, 1, Start),
              Regex("-", 1, Start)))
\end{verbatim}
}
\end{minipage}
\\
(a) & (b)
\end{tabular}
\caption{(a) The DSL $\mathcal{L}$ we consider for string transformation tasks. (b) An example synthesis task with four input-output examples together with one of the programs in the DSL that performs the string transformation required for transforming example inputs to corresponding outputs. }
\label{fig:dslexample}
\end{figure}

\textbf{RobustFill+Beam search.}
RobustFill~\citep{devlin2017robustfill} uses an encoder-decoder style architecture to learn to generate programs $P$ as a sequence of tokens in the DSL, conditioned on a set of input-output examples. Since the decoded programs can be tested at inference time to check whether they are consistent with the examples, it uses a beam search to generate the top-$K$ programs from the beam, and returns the first one (if any) that is consistent with the examples after execution.

\begin{figure*}
	\centering
    \includegraphics[width=\columnwidth]{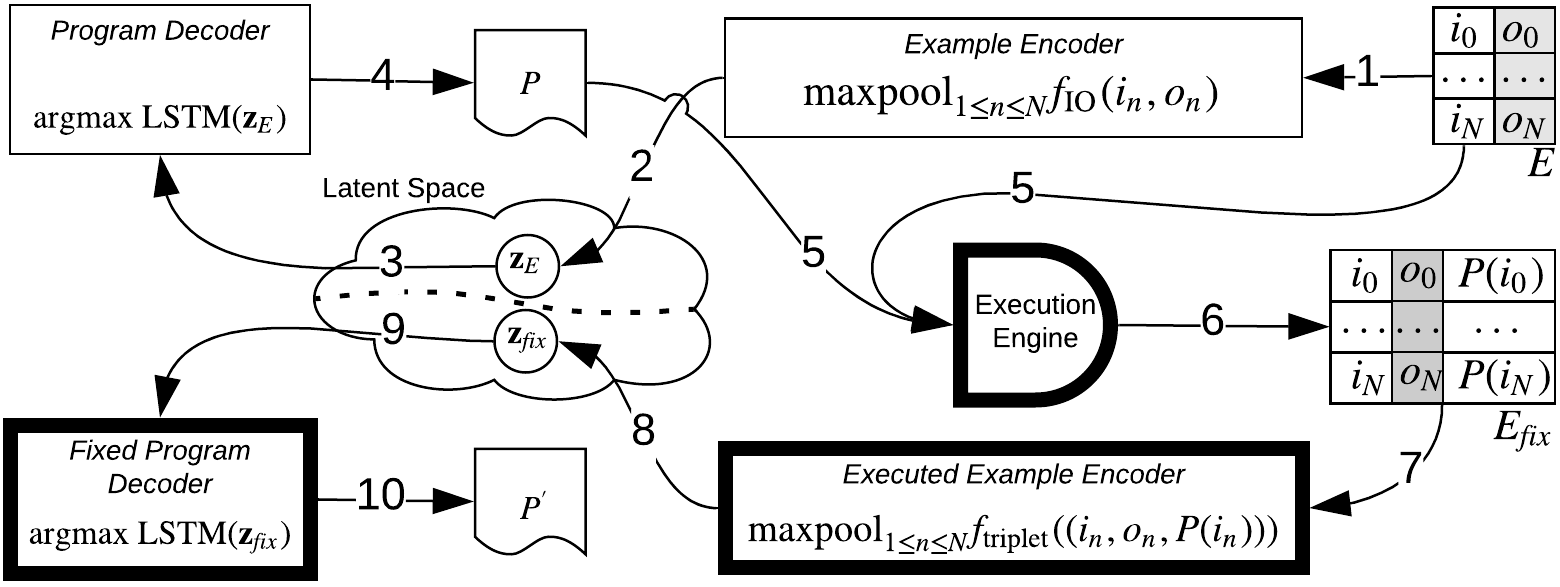}
    \caption{The high-level architecture for the system. We also study two variants: a) \emph{Shared Encoder} implements the \emph{Example Encoder} using  the \emph{Executed Example Encoder} with dummy $P(i_n)$, and b) \emph{Shared Decoder} uses the same decoder for both \emph{Program Decoder} and \emph{Fixed Program Decoder}. The \emph{differentiable fixer} consists of the components drawn with thick outlines.}
    \label{fig:system-design}
\end{figure*}

\subsection{Synthesis and Fixer Architecture}
\label{sec:modelarchitecture}

Taking inspiration from the fact that human developers rarely write the correct program on the first attempt, we introduce a new synthesis architecture with an additional neural component called the \emph{differentiable fixer}. The differentiable fixer takes the input-output examples, as well as the outputs generated by the execution of $P$ on the provided example inputs.
It then learns to perform corrections on $P$, to generate a ``better'' program $P'$ that can potentially correct the erroneous behavior of $P$ on the example inputs. Note that the fixer does not see the program $P$ itself, only its outputs on the example inputs; instead, it is given, in a sense, the chance to try again on what the first-round encoder/decoder attempted, with some extra information about how the first attempt turned out.

To illustrate the approach, consider the running example of Figure~\ref{fig:dslexample}b, and assume that the first time around, the program decoder produced the incorrect program $P$, with a faulty 3rd expression:

{\small\texttt{Concat(..., ..., SubStr(\underline{Regex(Word, 2, Start)}, Regex(",", 1, Start)), ..., ...)}}

This expression specifies the \emph{second} word of the input as the last name, rather than the \emph{last} word. Although this works just fine for examples 1, 3, and 4, it fails on example 2, which contains 3 words, and produces the incorrect output $P(i_2)=$ \texttt{"B. M : 617"}, instead of the correct $o_2=$\texttt{"B. Myers : 617"}.
The differentiable fixer takes as input the 4 \emph{triples} consisting of all input strings, desired output strings, and computed output strings -- note that for all but example 3, computed and desired outputs match -- and produces a "fixed" $P'$, which hopefully contains the correct clause shown in Figure~\ref{fig:dslexample}b.

The model architecture and the workflow are shown in Figure~\ref{fig:system-design}. There are three main components in our architecture: the \emph{example encoder}, the \emph{program decoder},
and the \emph{differentiable fixer},
which comprises an \emph{executed example encoder} and a \emph{fixed program decoder}. The first two components are the same as with the original baseline architecture; the differentiable fixer is the key new component introduced in this work.

The \textbf{example encoder} first embeds the input and output strings for each example, and then performs a max-pooling operation to obtain a representation of all the input-output examples 
\begin{equation*}
\mathbf{z}_E
=
\maxpool_{1 \leq n \leq N} f_{\text{IO}}( i_n, o_n )
,
\end{equation*}
where $f_{\text{IO}}$ is a learned network, and the max-pooling is applied elementwise.
In our implementation, $f_{\text{IO}}$ applies a learned character-level embedding to $i_n$ and $o_n$, concatenates all the resulting embeddings and passes them through 5 fully-connected layers.

The \textbf{program decoder} uses a recurrent neural network, specifically a long short-term memory network (LSTM), to generate a sequence of program tokens conditioned on the latent representation $\mathbf{z}_E$ obtained from the example encoder.
Abusing notation slightly, the program prediction can be written as $P = \operatorname{argmax} \operatorname{LSTM}(\mathbf{z}_E)$.
Note that a beam search can operate on top of the program decoder in order to decode multiple distinct program predictions.
Finally, the \textbf{differentiable fixer} takes as input the original input-output specification
and the outputs obtained from executing predicted program $P$, and outputs a fixed program $P'$.

The differentiable fixer  is implemented in a way that mirrors the original program synthesizer, and has three parts.
First, the \textbf{execution engine} defined by the DSL is used to
execute $P$ on the example inputs $i_0 \ldots i_N,$ yielding
predicted outputs $P(i_0) \ldots, P(i_N)$; note that the execution engine is part of the training procedure in our approach,
whereas it is only used at inference in the Robustfill approach.
Each of these predicted outputs
will be program values, e.g., strings in the RobustFill domain.
Then the \textbf{executed example encoder} maps the inputs, correct outputs,
and predicted outputs back to continuous representations.
It is similar to the example encoder, but also encodes the predicted
ouputs from executing $P$.
The example encoder produces a new \emph{fixed representation} of the program as
\begin{equation*}
\mathbf{z}_\mathit{fix}
=
\maxpool_{1 \leq n \leq N} f_{\text{triplet}}( (i_n, o_n, P(i_n)) )
.
\end{equation*}
The function $f_{\text{triplet}}$ is a neural network whose
input is a triple of input, correct output and predicted output, and consists of a character-level embedding layer followed by concatenation of all embeddings and 5 fully-connected layers.
Finally, 
the fixed representation is decoded by the \textbf{fixed program decoder},
which has a similar architecture as the program decoder, to
obtain the fixed program
\begin{equation*}
P'
=
\operatorname{argmax} \operatorname{LSTM}(\mathbf{z}_{\mathit{fix}})
.
\end{equation*}
This describes a family of program-fixer architectures.
We have considered several variants including:
1) \emph{Shared Encoder}: the two encoders $f_{\text{IO}}$ and $f_{\text{triplet}}$ share weights (in this case the initial $\mathbf{z}_E$ is computed by providing a dummy padding value in place of the executed example $P(i_n)$),
2) \emph{Shared Decoder}: the two decoders share weights, and
3) \emph{Program Encoder}: the intermediate program $P$ is included in the executed example encoder via a separate program-encoder component.
By sharing weights between the encoders and decoders we encourage $\mathbf{z}_E$ and $\mathbf{z}_\mathit{fix}$ to share the same latent space, and indeed we found that untying these weights would only increase the number of trainable parameters and not help performance.
All results in the Experiments section thus use the first two variants.
Although we have also explored the third variant to some extent (see Section~\ref{sec:discussion}), its full exploration is left to future work.

\subsection{Training procedure}
\label{sec:method:training}

The architecture is trained end-to-end using gradient-based minimization of a loss function comprising two terms: (1) a standard RobustFill decoding loss, which is the negative log-probability of the ground truth program $P^{*}$ under the \emph{program decoder} LSTM initialized with $\mathbf{z}_E$, and (2) a decoding loss of the \emph{fixed program decoder}, which is the negative log-probability of the ground truth program $P^{*}$ under the fixed program decoder LSTM initialized with $\mathbf{z}_\mathit{fix}$ (recall Figure~\ref{fig:system-design}).

Note that gradients do not flow through the execution engine; the execution engine serves as an on-the-fly data generator that appends predicted outputs $\{ P(i_n) \}_n$ to data points $(P^{*}, \{ (i_n, o_n) \}_n)$ in order to form the triplets that are encoded by the \emph{executed example encoder} (and then decoded by the \emph{fixed program decoder}).

\subsection{Iterative Fixing}
\label{sec:method:iterative}

At prediction time, the differentiable fixer can be applied iteratively to produce an $S$-best list of programs.
The previous fixed program $P'$ is then executed on the IO specification, and checked for correctness.
If the prediction does not match the input-output examples, the fixer is applied again to propose a new program $P''$.
To ensure that $P''$ is different from previous predictions,
beam search decoding is applied to the
fixed program decoder to return the first program that has previously not been predicted, comparing programs using syntactic equality.
This process is iterated until either a program is found that satisfies the IO examples, or a maximum number $S$ of iterations is reached.

\section{Experiments}
\label{sec:experiments}

We now present the experimental evaluation of our synthesizer with the differentiable fixer.
We explore the following research questions:
\begin{itemize}
\item
{\bf RQ1: On the RobustFill baseline, what is the effect of model size and program length?} Because the differentiable fixer increases the size of the model, to do a fair comparison we need to understand how a baseline {seq2seq} responds to increasing model size. We also examine how the baseline performance depends on the length of the ground-truth program, because for this task length is correlated with problem difficulty (Section~\ref{sec:experiments:baseline}).
\item
{\bf RQ2: Does the differentiable fixer improve accuracy?} Given the same budget of program executions (which determine if a program is correct), we demonstrate that utilizing our differentiable fixer improves performance on synthesis compared to the RobustFill baseline (Section~\ref{sec:experiments:search}).
\item
{\bf RQ3: How do the fixer predictions differ qualitatively from RobustFill?} We demonstrate the qualitative difference that the differentiable fixer makes, by analyzing the ``fixes'' it performs on initially incorrect programs, and comparing these to search dynamics in beam search (Section~\ref{sec:experiments:step2analysis}).
\end{itemize}

We perform experiments on a dataset generated by sampling programs from the DSL $\mathcal{L}$ (Figure~\ref{fig:dslexample}(a)) and the corresponding input-output examples using an invariant-based dataset generation method similar to the one used in NSPS~\citep{parisotto2017neuro} and RobustFill~\citep{devlin2017robustfill}.
The training and test datasets are made of \emph{tasks}, each of which contains a program $P^{*} \in \mathcal{L}$, and a set of four input-output examples.
The neural synthesizer takes as input the observed examples and generates a program $P$, which is then executed on the input-output examples to evaluate its correctness. We consider programs comprising up to $10$ expressions in the top-level \t{Concat} operator, and limit the input and output lengths to be of maximum 80 characters. The models are trained on 512 million tasks (seeing each once), and we evaluate the model performance on 1,000 held-out tasks.

\subsection{Effect of Model Size and Program Length on Seq2Seq Baseline}
\label{sec:experiments:baseline}

Our baseline is a seq2seq model without a fixer, that is, the example encoder and
the program decoder from Section~\ref{sec:method}. 
First we explore the performance of the baseline model with increasing model size and increasing program length to ensure that we are performing a fair
comparison.

\paragraph{Model size}

\begin{figure*}
	\centering
	\begin{subfigure}{.3\linewidth}
	    \centering
	    \includegraphics[width=\columnwidth]{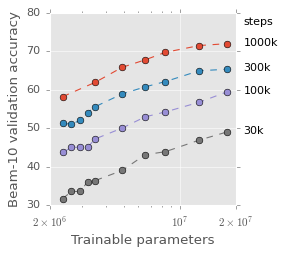}
		\caption{}
		\label{fig:robustfill_number_of_trainable_parameters}
	\end{subfigure}
 	\hfill
	\begin{subfigure}{.22\linewidth}
	    \footnotesize
	    \centering
		\begin{tabular}{cc}
            \toprule
            Length & Acc. \\
            \midrule
            $1$ & $93.2\%$ \\
            $2$ & $83.8\%$ \\
            $3$ & $71.1\%$ \\
            $4$ & $60.1\%$ \\
            $5$ & $49.1\%$ \\
            $6$ & $40.6\%$ \\
            $7$ & $33.2\%$ \\
            $8$ & $27.8\%$ \\
            $9$ & $22.7\%$ \\
            $10$ & $16.6\%$ \\
            \bottomrule
        \end{tabular}
        \caption{}
	    \label{tab:length_baseline_accuracy}
	\end{subfigure}
 	\hfill
	\begin{subfigure}{.37\linewidth}
	    \centering
	    \footnotesize
	\begin{tabular}{cccccc}
        \toprule
        Algorithm & $H$ & Params & Acc. \\
        \midrule
        Beam & $512$ & $2.2M$ & $53\%$ \\
        Beam & $576$ & $2.9M$ & $54\%$ \\
        Beam & $608$ & $3.2M$ & $56\%$ \\
        Fixer & $512$ & $3.0M$ & ${\bf 61}\%$        \\
        \bottomrule
    \end{tabular}
        \caption{}
	    \label{tab:step10_accuracy}
	\end{subfigure}
	\caption{\small
	    \emph{(a)}: Influence of hidden size on baseline accuracy.
	    \emph{(b)}: Baseline accuracy by ground truth program length.
	    \emph{(c)}: Baseline vs Differentiable Fixer accuracies on the full validation data after $10$ steps of search.
	}
\end{figure*}

Because the differentiable fixer model is an extended, larger version of the baseline model, a natural question is whether the performance of the baseline can be  
more simply enhanced just by scaling up the model size.
Indeed, we observed that the baseline seq2seq model responds well to having its hidden dimension increased.
Figure~\ref{fig:robustfill_number_of_trainable_parameters} shows that as a rule of thumb, doubling the number of trainable parameters leads to a $5$ percentage point improvement in validation accuracy performance.
(Note that the number of trainable parameters is asymptotically quadratic in the hidden size $H$.)
Additionally, with a hidden size $H = 512$ and beam size $10$, we match or surpass the $50\%$ validation accuracy of the ``Basic Seq.'' model reported in~\citet[App. B]{devlin2017robustfill}, when accounting for the number of data points seen in training.
As a result, we make sure to normalize for model size in the comparisons
in the next section.

\paragraph{Impact of program length}

We examine whether the seq2seq model performs worse at synthesizing longer programs,
to see if this is a useful proxy measure of problem difficulty when comparing models.
We define the \emph{length} of a RobustFill program as the number of \texttt{Expression}s that appear in the top-level \texttt{Concat} operation.
Figure~\ref{tab:length_baseline_accuracy} shows how the accuracy of the baseline RobustFill model decreases with the complexity of the problem, as measured by the ground-truth program length.
As expected, the higher the program length, the more challenging the synthesis task: we observe a gradual decrease in the model accuracy with increasing program lengths. This will be important in contextualizing the subsequent analysis, where the differentiable fixer provides improvements on harder tasks (up to $20\%$ more solved instances).

\subsection{Synthesizer Accuracy with Differentiable Fixer vs Beam Search}
\label{sec:experiments:search}

In this section, we analyze the performance improvement due to the differentiable fixer.
In particular, we compare two such combinations of model and search procedures:
\begin{enumerate}
\item
\emph{Baseline (seq2seq + beam search)}:
Given a set of IO examples, the seq2seq model encodes them, and then a beam search is run on the decoder, with beam size $S = 10$.
This guarantees that we obtain $S$ \emph{unique} candidate programs.
We successively execute each of the candidate programs on the inputs in the provided IO examples, starting from the most likely program, and stop with success whenever the generated program outputs match exactly the example outputs.
Note that there are (at most) $S$ candidate programs executed.
\item
\emph{Differentiable fixer}:
The differentiable fixer is applied iteratively, as described in Section~\ref{sec:method:iterative}, for at most $S$ steps.
\end{enumerate}

All models use an identical batch size $B = 1024$, and are trained for $500{,}000$ steps.
We evaluate the methods by accuracy, that is, the proportion of the validation tasks on which
any of the $S$ steps of search satisfies all of the input-output examples.

Figure~\ref{tab:step10_accuracy} reports the accuracies of both methods.
The baseline model with hidden size $H = 512$ achieves an accuracy of $53\%$.
Using a larger model with hidden size $H = 608$ (resulting in approximately 3.2~M trainable parameters), the model accuracy increases by $3$ percentage points to $56\%$.
On the other hand, using our differentiable fixer in the original baseline model results in an accuracy gain of $8$ percentage points to $61\%$.
Controlling for the same number of trainable parameters, the differentiable fixer achieves a $5$-percentage-point advantage over simply increasing the baseline model size.

\subsection{Qualitative Analysis of Fixes}
\label{sec:experiments:step2analysis}

In order to understand the qualitative differences between the methods, we perform an analysis of cases where each method made a successful correction at step 2 after an initially incorrect prediction at step 1.
To this end, we sampled $100{,}000$ evaluation tasks with ground-truth program length $10$.
(Appendix~\ref{app:step2} repeats the analysis on all program lengths.)
This is to eliminate the effect of different ground-truth program length on the analysis, and also to focus on the hardest instances.
For both the beam search baseline and the differentiable fixer we consider the cases where the initial prediction was incorrect, and the second prediction was correct.
This is so that we can analyse the kinds of corrections each method makes when the top prediction is incorrect.

Overall, beam search made $2{,}309$ successful corrections at step 2, while the fixer made $5{,}367$.
Note that since the RobustFill baseline accuracy at step $1$ is only $17\%$ for this challenging subset of programs (Figure~\ref{tab:length_baseline_accuracy}), the fixer is providing a more substantial performance improvement here relative to the baseline accuracy.
Putting aside the increased performance of the fixer, we seek to understand the different properties of these fixes next.

\begin{figure*}[ht]
	\centering
	\begin{subfigure}{.31\linewidth}
	    \centering
		\includegraphics[width=\columnwidth]{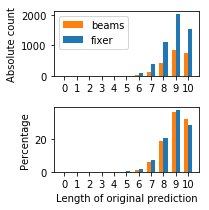}
		\caption{}
		\label{fig:step2:length10:original_prediction_length}
	\end{subfigure}
	\hfill
	\begin{subfigure}{.31\linewidth}
	    \centering
		\includegraphics[width=\columnwidth]{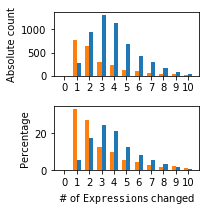}
		\caption{}
		\label{fig:step2:length10:expressions_changed}
	\end{subfigure}
	\hfill
	\begin{subfigure}{.31\linewidth}
	    \centering
		\includegraphics[width=\columnwidth]{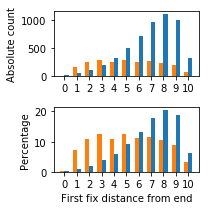}
		\caption{}
		\label{fig:step2:length10:distance_from_end}
	\end{subfigure}
	\caption{\small{
	    Successful corrections at step 2 of the search using beam search and the differentiable fixer. Ground truth program length is 10.
	    \emph{Left}: Length of the original prediction at step $1$.
	    \emph{Middle}: Number of \texttt{Expression}s changed at step $2$.
	    \emph{Right}: Distance of earliest fix from the end of the program.
	}}
	\label{fig:step2:length10}
\end{figure*}

{\bf Distribution of corrections by original prediction complexity. }Figure~\ref{fig:step2:length10:original_prediction_length} shows the number of successful corrections stratified by the length of the original incorrect prediction (the number of \texttt{Expression}s in the top-level \texttt{Concat}).
The top subfigure shows absolute counts, and confirms that for all lengths of the original prediction the fixer is able to perform more successful corrections than beam search.
The second subfigure hints that disproportionately more Beam successes come on predictions that had length $10$ to start with, whereas the fixer is more often able to fix predictions that were perhaps too short to start with.
The analysis across different program lengths in Appendix~\ref{app:step2} supports the claim that the fixer is able to lengthen original predictions that were just too short, whereas the Beam almost never performs that type of correction successfully.

{\bf Distribution of Complexity of Corrections. } Figure~\ref{fig:step2:length10:expressions_changed} analyzes the number of {expression}s in the top-level \texttt{Concat} that are changed between the initial incorrect prediction and the fixed program.
The normalized counts in the bottom subfigure reveal the qualitative difference that, whereas beam search only changes a single \texttt{Expression} in more than $30\%$ of cases, the fixer tends to change three, four, or more \texttt{Expression}s simultaneously more often than beam search.

We confirmed that the observed fix distribution looks identical when the beam search is executed on top of a model trained with hidden size $H = 1024$, i.e.,~with~$\sim 4$x as many trainable parameters (more than the parameters in the differentiable fixer).
This confirms that the qualitatively different shape of the fixer's distribution cannot be simply attributed to it being a larger model.

{\bf Distribution of Correction Locations. } Finally, we investigate the hypothesis that beam search is more prone to path degeneracy, i.e., that elements of the beam tend to share long common prefixes and only differ in shorter suffixes.
To this end, we analyze the furthest distance \emph{from the end of the program} where an edit of a successful correction takes place.
A value of $1$ means that the last {expression} in the top-level \texttt{Concat} of the initial (incorrect) prediction was modified (but nothing earlier).
A value of $0$ means that the original prediction was left intact, but a new {expression} was \emph{inserted}, i.e.,~that the corrected prediction has one more \texttt{Expression} than the initial (incorrect) one.

The absolute numbers in Figure~\ref{fig:step2:length10:distance_from_end} show that the fixer performs more successful corrections at larger distances from the end of the program, even though beam search is also surprisingly strong at making changes far away from the end of the program.
Further, the analysis in Appendix~\ref{app:step2} shows that the fixer exhibits the ability to lengthen the initial (correct) prediction while beam search almost never performs a correction of this kind (see column $0$ in Figure~\ref{fig:step2:lengthall:distance_from_end}).
Together, these observations substantiate our claim that the differentiable fixer is able to predict qualitatively more complex corrections than beam search.

\section{Conclusion and Future Work}
\label{sec:discussion}

We proposed a neural architecture for program synthesis augmented with a \emph{differentiable fixer}, a module that can be used to fix initially incorrect predictions.
We showed that adding the differentiable fixer to a baseline seq2seq model is a more efficient way of increasing the model size, as it provides an inductive bias that leads to higher validation accuracies.
We also analyzed the qualitative difference between fixing incorrect predictions and predicting multiple candidate programs from a beam search, demonstrating that the learned fixer is more effective.

We have additionally explored extensions to the family of models based on differentiable program fixer; although our initial investigations did not show conclusive improvements and more thorough analysis is left to future work. First, we have introduced a \emph{learned program encoder}, which provides the fixer with a representation of the (failed) candidate program, in addition to that program's outputs. Our RNN-based program encoder did not seem justified as using the same additional parameter budget on the fixer itself yielded better results. This variant could possibly be improved by using program encoders based on grammar or static-analysis graphs. Second, we tried incorporating several types of auxiliary losses during training: predicting program length, program grammatical validity, per-concat expression validity, and per-example correctness, but none showed marked improvement.

We further hope to investigate how a differentiable fixer interacts with more sophisticated baseline architectures, such as attention modules and Transformers.
Finally, an exciting direction for future work is to combine our approach with
execution-guided synthesis, where intermediate steps during the construction of a candidate program are informed by partial execution traces, especially for languages in which partial programs can be meaningfully executed, even beyond string processing.

\begin{ack}
The authors would like to gratefully acknowledge helpful comments by Hanjun Dai, Hyeontaek Lim, and Rif A. Saurous.
\end{ack}

\small
\bibliographystyle{abbrvnat}
\bibliography{bibliography}

\newpage
\normalsize
\appendix

\section{Step 2 analysis across all program lengths}
\label{app:step2}

Here we repeat the analysis from Section~\ref{sec:experiments:step2analysis} but instead of sampling $100{,}000$ data points with equal program length $10$, here we sample $100{,}000$ data points from the full training distribution, i.e.~containing programs of all lengths up to $10$.
As before We ran beam search and fixer search for 2 steps on top of the same trained model and for both methods individually we eliminated cases where the initial prediction was already a correct solution, or where neither the first nor the second prediction solved the problem.
We were thus left with only those cases where a successful correction of an initially incorrect prediction was made at step $2$ of the search.

Overall, beam search made $3{,}642$ such corrections, while the fixer made $7{,}935$.
Putting aside the increased performance of the fixer, we seek to understand the different properties of these fixes next.

\begin{figure*}[ht]
	\centering
	\begin{subfigure}{.31\linewidth}
	    \centering
		\includegraphics[width=\columnwidth]{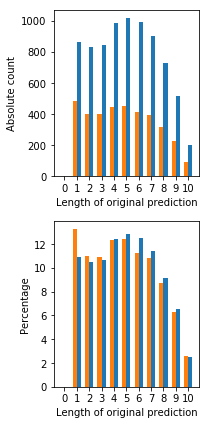}
		\caption{}
		\label{fig:step2:lengthall:original_prediction_length}
	\end{subfigure}
	\hfill
	\begin{subfigure}{.31\linewidth}
	    \centering
		\includegraphics[width=\columnwidth]{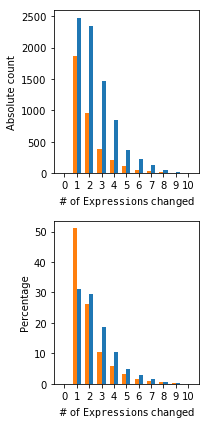}
		\caption{}
		\label{fig:step2:lengthall:expressions_changed}
	\end{subfigure}
	\hfill
	\begin{subfigure}{.31\linewidth}
	    \centering
		\includegraphics[width=\columnwidth]{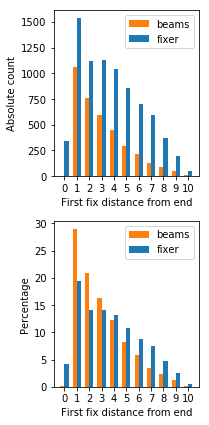}
		\caption{}
		\label{fig:step2:lengthall:distance_from_end}
	\end{subfigure}
	\caption{\small{
	    Successful corrections at step 2 of the search using beam search and the differentiable fixer.
	    \emph{Left}: Original prediction length.
	    \emph{Middle}: Number of expressions changed.
	    \emph{Right}: Distance from end.
	}}
	\label{fig:step2:lengthall}
\end{figure*}

\paragraph{Distribution of corrections by Program complexity}

Figure~\ref{fig:step2:lengthall:original_prediction_length} shows the number of successful corrections stratified by the length of the original incorrect prediction (the number of \texttt{Expression}s in the top-level \texttt{Concat}).
The first subfigure shows absolute counts, and confirms that for all program lengths the fixer is able to perform more successful corrections than beam search.
Interestingly, the second subfigure shows that disproportionately more Beam successes come on a program of length only 1, meaning that disproportionately more Fixer successes are on the longer, more difficult examples.

\paragraph{Distribution of complexity of Corrections}

Figure~\ref{fig:step2:lengthall:expressions_changed} analyzes the number of \texttt{Expression}s in the top-level \texttt{Concat} that are changed between the initial (incorrect) prediction and the correct program discovered in step 2 of the search.
The absolute numbers in the first subfigure confirm that the fixer performs at least as many successful fixes of a given size as the beam.
However, the normalized counts reveal the qualitative difference that, whereas beam search only changes a single \texttt{Expression} in more than $40\%$ of cases, the fixer tends to change two, three, or four \texttt{Expression}s simultaneously more often than beam search.

We confirmed that the observed fix distribution looks identical when the beam search is executed on top of a model trained with hidden size $H = 1024$, i.e.,~with~$\sim 4$x as many trainable parameters (more than the parameters in the Differentiable Fixer).
This confirms that the qualitatively different shape of the fixer's distribution cannot be simply attributed to it being a larger model (i.e.,~having more trainable parameters).

\paragraph{Distribution of Correction Locations}

Finally, we investigate the hypothesis that beam search is more prone to path degeneracy, i.e.~that elements of the beam tend to share long common prefixes and only differ in shorter suffixes.
To this end, we analyze how far \emph{from the end of the program} the first edit of a successful correction takes place.
A value of $1$ means that the last \texttt{Expression} in the top-level \texttt{Concat} of the initial (incorrect) prediction was modified.
A value of $0$ means that a new \texttt{Expression} was \emph{inserted}, i.e.,~that the corrected prediction has one more \texttt{Expression} than the initial (incorrect) one.

The absolute numbers in Figure~\ref{fig:step2:lengthall:distance_from_end} show that the fixer performs more successful corrections at all distances from the end of the program, even though beam search is also surprisingly strong at making changes far away from the end of the program.
However, strikingly, the fixer exhibits the ability to lengthen the initial (correct) prediction while beam search almost never performs a correction of this kind (see column $0$).

The analysis above substantiates our claim that the differentiable fixer is able to predict qualitatively more complex corrections than beam search; we have seen that not only are multiple \texttt{Expressions} edited simultaneously more often, but the differentiable fixer is also much more effective in changing the overall length of the initial prediction.

\end{document}